\documentclass[10pt,twocolumn,letterpaper]{article}

\usepackage{cvpr}
\usepackage{times}
\usepackage{epsfig}
\usepackage{graphicx}
\usepackage{amsmath}
\usepackage{amssymb}

\usepackage{mathtools}
\usepackage{algorithm}
\usepackage{algorithmic}
\usepackage{mathrsfs}
\usepackage{amsfonts}

\usepackage{cite}
\usepackage{subeqnarray}
\usepackage{multirow}
\usepackage{epsfig}
\usepackage{mathptmx}
\usepackage{times}
\usepackage{booktabs}
\usepackage{sansmath}

\usepackage[pagebackref=true,breaklinks=true,letterpaper=true,colorlinks,bookmarks=false]{hyperref}

 \cvprfinalcopy 

\ifcvprfinal\fi
\begin{document}

\title{AprilTags 3D: Dynamic Fiducial Markers for Robust Pose Estimation in Highly Reflective Environments and  Indirect Communication in Swarm Robotics}

\author{Luis A. Mateos\\
MIT\\
}

\maketitle

\begin{abstract}
   Although fiducial markers give an accurate pose estimation in laboratory conditions, where the noisy factors are controlled, using them in field robotic applications remains a challenge. This is constrained to the fiducial maker systems, since they only work within the RGB image space. As a result, noises in the image produce large pose estimation errors. In robotic applications, fiducial markers  have been mainly used in its original and simple form, as a plane in a printed paper sheet. This setup is sufficient for basic visual servoing and augmented reality applications, but not for complex swarm robotic applications in which the setup consists of multiple dynamic markers (tags displayed on LCD screen).
   
   This paper describes a novel methodology, called AprilTags3D, that improves pose estimation accuracy of AprilTags in field robotics with only RGB sensor by adding a third dimension to the marker detector. Also, presents experimental results from applying the proposed methodology to swarm autonomous robotic boats for latching between them and for creating robotic formations.
   
\end{abstract}

\section{Introduction}

\begin{figure}[t]
	\begin{center}
		\includegraphics[width=0.95\linewidth]{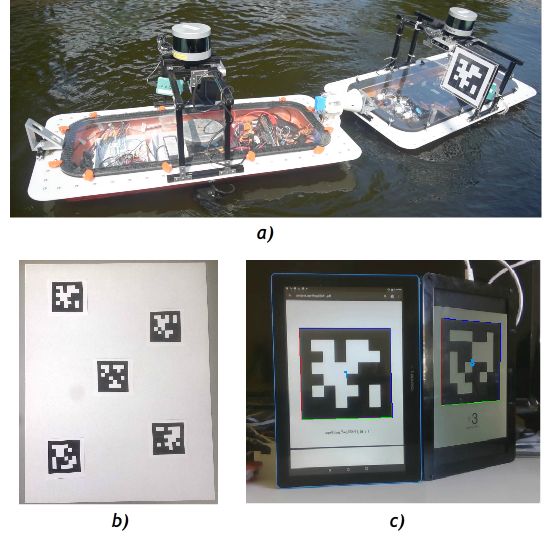}
	\end{center}
	\caption{A) Classic AprilTag printed in paper sheet used for pose estimation in autonomous latching with robotic boats. B) Tag bundle setup with five markers for more accurate pose estimation (the markers are set in the same plane). C) AprilTags3D setup with a couple of markers rotated each other. The markers are displayed on LCD screens and are able to change their tagID to indirectly communicate their state to other robots without losing their position and orientation. }
	\label{fig:1}
\end{figure}

Fiducial markers are commonly use in computer vision (CV) and augmented reality (AR) applications for detection and identification. In robotic applications, fiducial markers have been of crucial importance for obtaining an accurate pose estimation of the marker. Since, estimating the tag's 6 Degrees of Freedom (DOF) can lead a robotic arm to grab an item, or guide an autonomous robotic boat to a docking station or to latch another robot \cite{latching}, see Figure \ref{fig:1}. 

There are different families of markers, with circular and squared shapes, and color base \cite{DeGol2017ChromaTagAC}. 
Circular tags such as Intersense \cite{1115065} and Rune tags \cite{5995544} provide accurate pose estimation in short distances with a high computational cost. While, squared fiducial tags such as ARTags \cite{fiala}, ARToolkit \cite{kato}, ArUco \cite{GARRIDOJURADO20142280}, AprilTags \cite{apriltags} and AprilTag2 \cite{apriltags2}  have a low computational cost and can be detected from a further distance. 

In robotics, the AprilTags framework has been preferred for having a lower false positive rate and higher detection rates, even in challenging viewing angles at further distances. However, AprilTags marker systems rely on RGB image space for detection and pose estimation, which is susceptible to the perspective ambiguity noise from sub-pixel detection in their corner locations. Resulting in rotational errors, making the pose estimation challenging without additional information \cite{8206468}.

\begin{figure}[t]
	\begin{center}
		\includegraphics[width=0.99\linewidth]{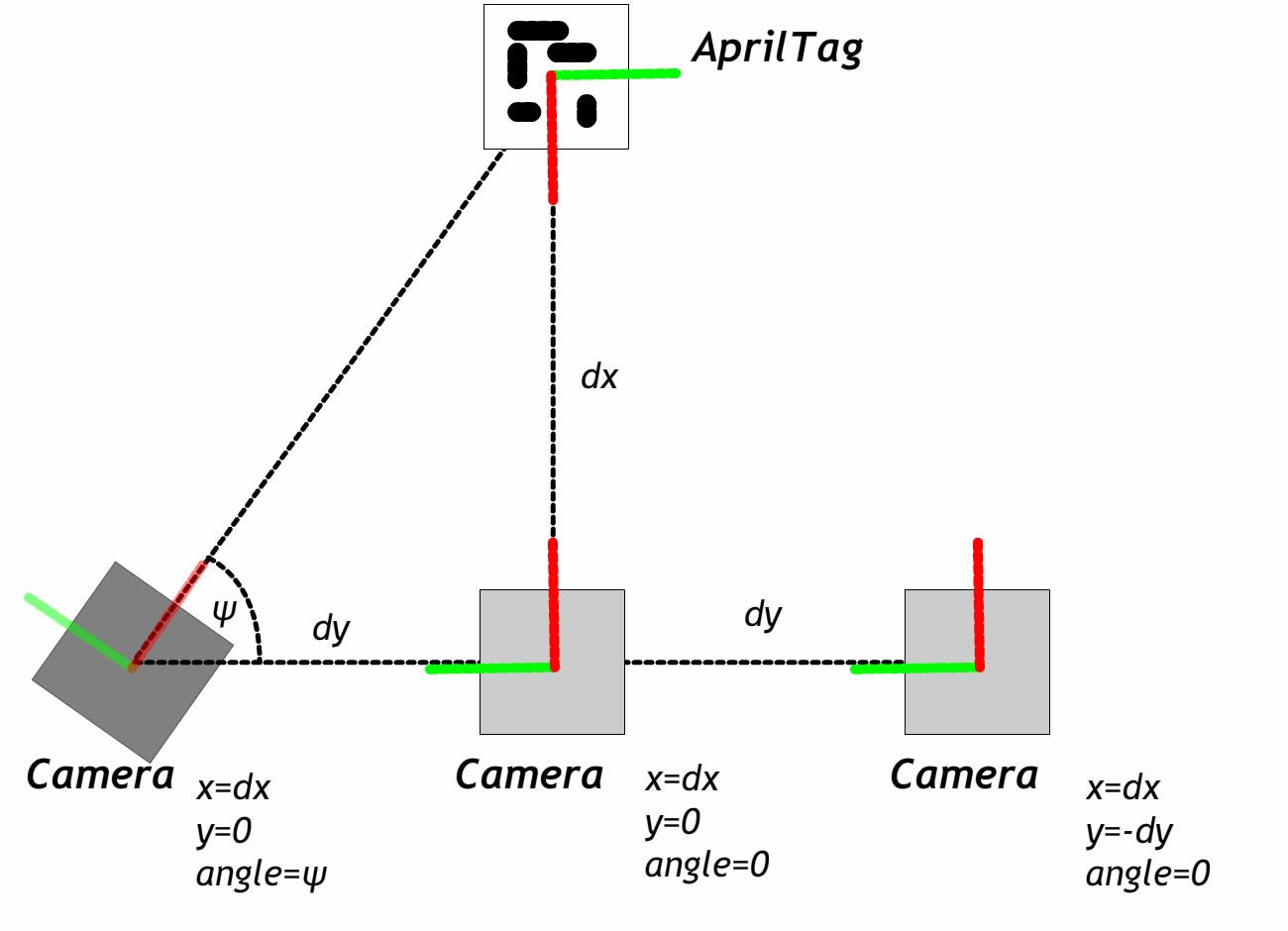}
	\end{center}
	\caption{AprilTags framework.}
	\label{fig:angle}
\end{figure}

Sensor fusion methods integrate additional sensors to cope with the RGB detection problems, such as RGBD, which correlates the markers pose detection with the plane they are positioned \cite{6506990}. Other approaches in the Simultaneous Localization and Mapping (SLAM) community propose fusing RGB with inertial data using extended Kalman filters to enhance the tracking algorithms accuracy \cite{7266817}. However, for our robotic setup we can only rely on RGB camera for pose estimation without additional sensor information.

In laboratory conditions where the light is regulated, a single tag is able to give accurate position and orientation. However, in field robotic applications with multiple light sources and reflections, relying on one tag will result in noisy and inaccurate estimations. 
In order to improve the pose estimation in noisy environments, a collection of AprilTags can be used simultaneously to extract a single "bundle pose", rather than the poses of the individual tags, see Figure \ref{fig:1}$b$. This is helpful since bundle pose estimation makes use of every detected tags' corners (making for $4n$ points if $n$ tags are detected) rather than single tag pose estimation, which only uses 4 points corresponding to the 4 tag corners \cite{7881437}.  However, the tags are set on the same plane and a reflection affecting one of them may affect all of them, and still resulting in a noisy pose estimation.

In order to obtain a reliable pose estimation with AprilTags using only RGB camera, we propose a simple rather powerful method, called AprilTags3D, which integrates two or more tags with the constraint that they are not laying in the same plane, see Figure  \ref{fig:1}$c$. Instead, the tags are rotated each other, so that in highly reflective environments at least one tag can still be detected and a proper pose estimation can be computed. The AprilTags3D can be described as a tag bundle where each tag is set on its own unique plane.

We present two main contributions in this paper. First, we describe a novel methodology to improve the pose estimation when only relying on RGB cameras. The methodology takes advantage of the third dimension by computing the 6 DOF from a couple or more markers, shifted in different connected planes and rotated a few degrees from each other, creating a 3D object (similar to a "half-disco ball"). 
In this way, the arrangement of the rotated tags improves the position estimation, by interpolating the position and orientation of the tags with a degree of reliability, mimicking a Kalman filter for sensor fusion, resulting in a robust position estimation in noisy conditions in real time. 

Also, we propose a method for dynamic identification and indirect communication in swarm robotic applications. The method, instead of incorporating the classic paper printed AprilTags on each robot, integrates LCD screens that change the displayed tag depending on the state of the robot without losing its position and orientation. 


\begin{figure*}[t]
	\begin{center}
		\includegraphics[width=0.99\linewidth]{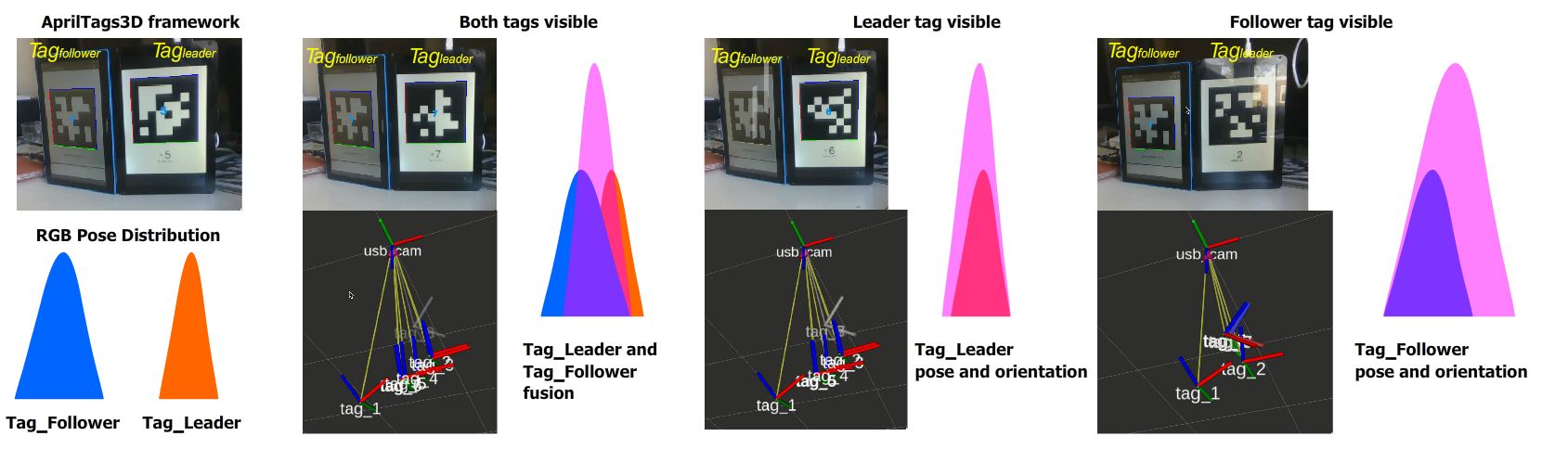}
	\end{center}
	\caption{An overview of our proposed method pipeline. Pose estimation from a couple of dynamic markers. The observations are combined  according to their uncertainty distribution similar to a Kalman filter for sensor fusion.}
	\label{fig:2}
\end{figure*}


\section{AprilTags3D}
Among squared fiducial marker systems, AprilTags \cite{apriltags} \cite{apriltags2} perform the best when detecting smaller markers, in high angle inclination and in different lighting levels \cite{7271491}. This is the reason of their popularity in the robotics community.

Figure \ref{fig:angle}, shows the AprilTags framework. If the camera is positioned in front of the tag, the lateral distance $d_y$ and angle $\psi$ are zero. While the longitudinal distance between the camera and tag is $d_x$ in the $X-axis$. If the camera changes its position with the same orientation, then the lateral distance $d_y$ reflects its change in position on the $Y-axis$, $d_x$ is the same longitudinal distance and  $\psi=0$. If the camera in that position changes its orientation to face the tag, then the lateral distance $d_y=0$ and the angle reading becomes the true angle $\psi$ between the camera and tag.

In AprilTags the error in position and orientation varies with the distance and rotation between the camera and the marker. The error in distance is linear, since the error increases the further away the marker is from the camera. 
The work from \cite{8278082} and \cite{Pentenrieder2007AnalysisOT} summarizes and quantifies the accuracy of marker-based 3D visual localization.





\subsection{AprilTags3D framework}
A  robotic setup integrating cameras and  sensors may be validated in laboratory conditions, however, this may fail when used outdoors in real world conditions.  This is due to the noises from the environment that cannot be fully tested in indoor setups, such as sunlight illumination and reflections of light from other objects. 

The aim of AprilTags3D framework is to minimize the error in markers detection and pose estimation in real case scenarios. The framework consists of two or more tags that are not lying on the same plane, as they are rotated but still linked to each other, preventing that a single light source distort or even cancel the tag detection and pose estimation.

The setup consists of at least a couple of markers, one leader marker $Tag_{leader}$ and one or more follower marker(s) $Tag_{follower\#}$. The markers are linked on one axis as if connected by a hinge and rotated certain degrees in a hyperbolic fashion.  
The degree of rotation is set from the distance and angle study to obtain the best detection rates \cite{8278082}  \cite{Pentenrieder2007AnalysisOT}. Further, the degree of rotation can be set upon the application and knowledge of the working environment of the robot. 

 Hence, the tags are no longer forming a plane, as they become a 3D object. This novel setup enables us to keep the pose estimation even if the environment is  extremely noisy. Moreover, we integrate LCD screens to display the tags dynamically, overcoming low light conditions and enabling indirect communication between robots. 
 In this sense, a robot in a swarm can notify other robots about its state by changing the tag displayed on its LCD without affecting the tag detection, neither the pose estimation, see Figure \ref{fig:2}.

\subsubsection{AprilTags3D markers} 
The simplest configuration integrates a couple of markers, a $Tag_{leader}$ and a $Tag_{follower}$ connected in the $Z-axis$ and rotated $g$ degrees. 
The marker nearer the camera will receive a higher weight, since the error is lower in short distances \cite{8278082}. The marker with the best orientation will receive another weight   \cite{Pentenrieder2007AnalysisOT}. The process is to combine the weights and take the higher confidence level or "heavier".


In the optimal case, when the two tags are detected, the data is filtered with a nonlinear digital filtering technique (median) and then the joint pose estimation is computed from the couple of 6 DOF to refine the real pose estimation. 
For example, Figure \ref{fig:2} shows the physical setup consisting of a couple of LCD screens, $Tag_{leader}$ linked to $Tag_{follower}$ rotated $10^\circ$ in the Z -axis. As long as the $Tag_{leader}$ is visible and nearer to the camera its pose distribution is more reliable than the $Tag_{follower}$, since the $follower$ is rotated and with larger distance to the camera. 
In the case the $Tag_{leader}$ is the only one detected, we rely on it similar to the classic single marker framework. While, if the $Tag_{follower}$ is the only one detected, it is taken into account as a single marker with its corresponding confidence degree.

\begin{figure}[b]
	\begin{center}
		\includegraphics[width=0.8\linewidth]{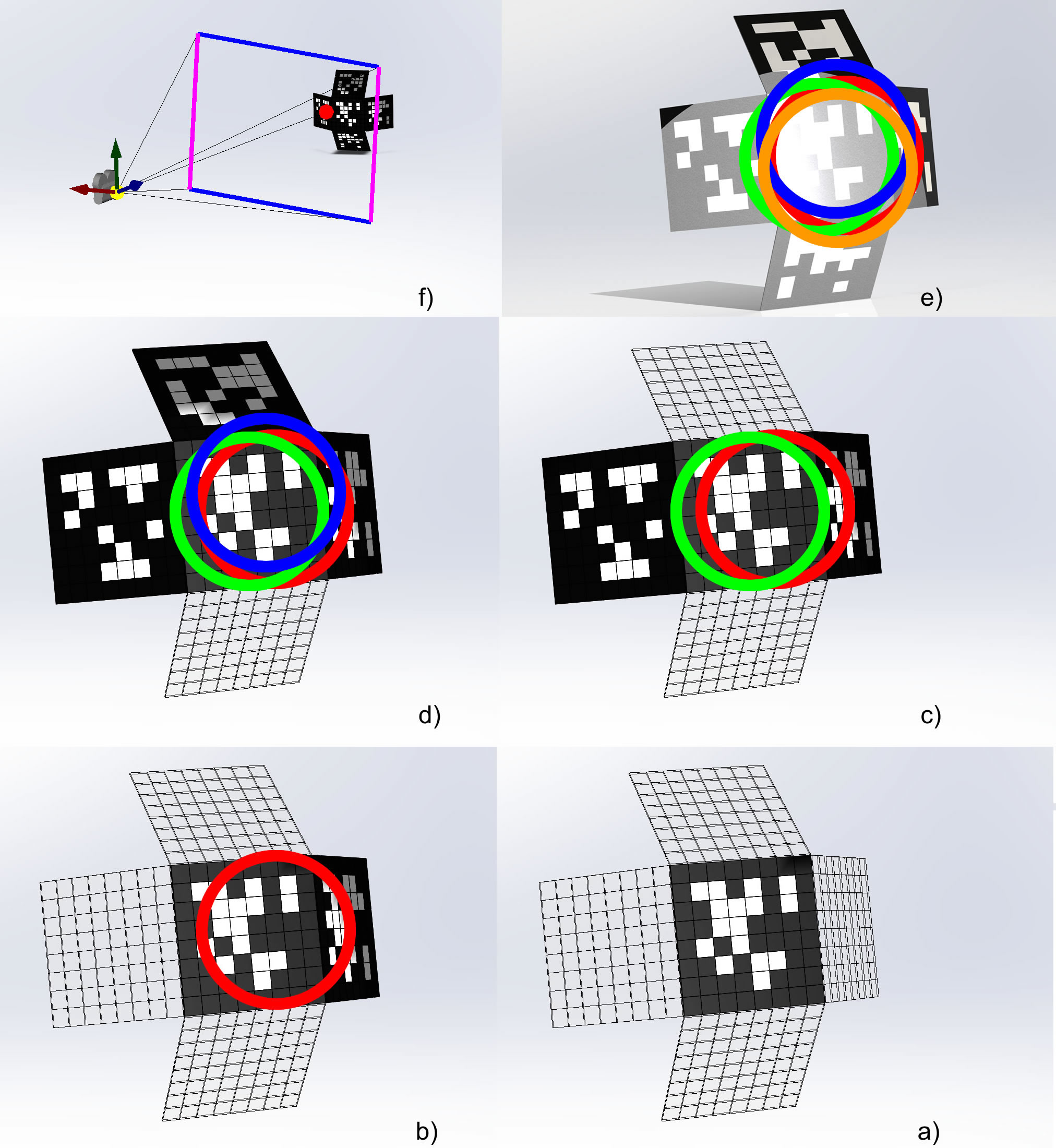}
	\end{center}
	\caption{AprilTags 3D. a) Undetectable tag due to direct light reflection. b) Additional tag is detected. c) Tag refines pose. d) Tag improves pose estimation. e) The more the measurement signals become available, the estimation will be more accurate. }
	\label{fig:kalman}
\end{figure}

The framework is similar to a Kalman filter for sensor fusion. In this case, instead having sensors there are  markers, and the noise from the sensors in the filter is defined as the degree of confidence from the rotation and distance of the markers to the camera. In this  scheme, the advantage is that although only one measurement is available, the pose estimation can be done. The more the measurement signals become available, the estimation will be more accurate, see Figure \ref{fig:kalman}.
 
\begin{figure*}[t]
	\begin{center}
		\includegraphics[width=0.99\linewidth]{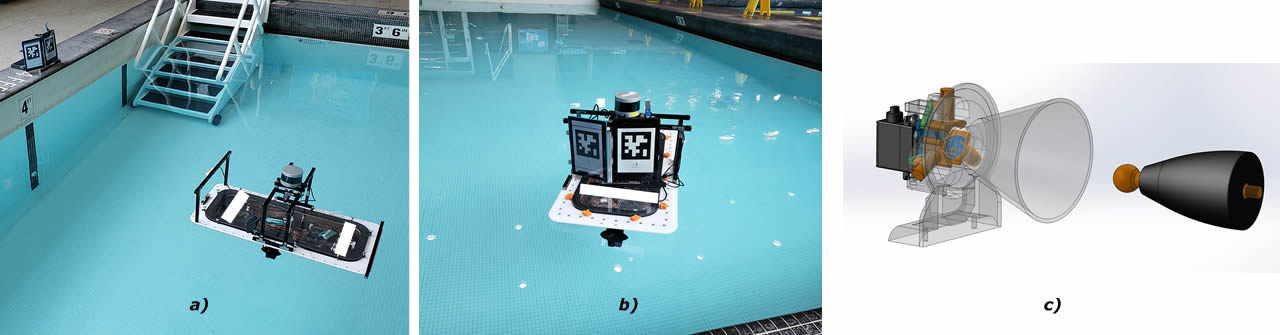}

	\end{center}
	\caption{A) Autonomous robotic boat guided to docking with AprilTags3D. B) AprilTags3D with a couple of LCD screens on the robotic boat for dynamic latching and indirect communication. C) Latching system with actuated funnel and pin with bearing stud.}
	\label{fig:roboat}
\end{figure*}

\subsubsection{AprilTags3D for Indirect Communication in Swarm Robotics}
Swarm robotics refer to robots controlled by them-self, without a centralized control coordinator. If individual robots integrate the AprilTags3D framework with LCD screens and cameras. Then, these independent entities can indirectly communicate between them by been able to detect the tag on another robot and infer the meaning of its marker. 

If a robot with a camera $R_c$ is following the tag on another robot $R_t$ and the tag is dynamically changing. The pose estimation computed by $R_c$ will not be affected by the changing tag. Instead, $R_c$ reads a different tagID that may communicate a changing state in $R_t$ or a direct order to $R_c$ to perform certain action.  
This change of tagID can be generalized for swarm robotics as one robot can communicate with all other robots or with only one is specific. 

Moreover, the complexity of this indirect communication is minimal and can be extended to $N$ robots, as long as each robot can see the tag on another robot. Section \ref{swarms} shows experimental results of  AprilTags3D for robot formations.


\begin{figure}[b]
	\begin{center}
		\includegraphics[width=0.9\linewidth]{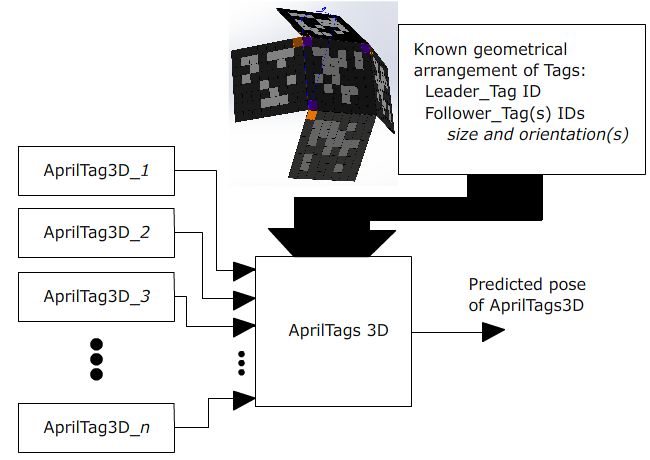}
	\end{center}
	\caption{
		The fused signal is more accurate than any of the original signals if considered separately. The enhanced, more accurate signal is used to guide and navigate the robot.}
	\label{fig:controller}
\end{figure}

\section{Autonomous Robotic Boat}
The robotic boat consists of a rectangular base (2:1 ratio) with four thrusters in the middle of the edges, see Figure \ref{fig:roboat}$a$. 
In this relationship the robot is able to move forward, backward, sideways and able to rotate on its axis. The dimension of the robotic platform are $1000mm\times500mm\times150mm$. 
The dynamics and model of the robotic boat can be found in \cite{weix}. 

The robotic boat integrates a $360^\circ$ lidar for navigation and obstacle avoidance for moving long distances $D>10m$. While, for latching in short distances $d<10m$ the camera - target principle is implemented.

\subsection{Latching system}
The robotic boats are able to latch to a docking station or to another boat with a couple of mechanisms: 1) an actuated funnel and 2) a pin with a bearing stud, see Figure \ref{fig:roboat}$c$. In both cases the robot detects with its camera the tag on the other entity, computes the pose estimation and latch to the pin on the robot with the tag or to the docking station \cite{latching}. 

When the latching is achieved, the funnel and pin creates a spherical joint between the entities that enables them to overcome wave disturbances while connected.

\subsubsection{Guiding system}
\begin{figure}[b]
	\begin{center}
		\includegraphics[width=0.9\linewidth]{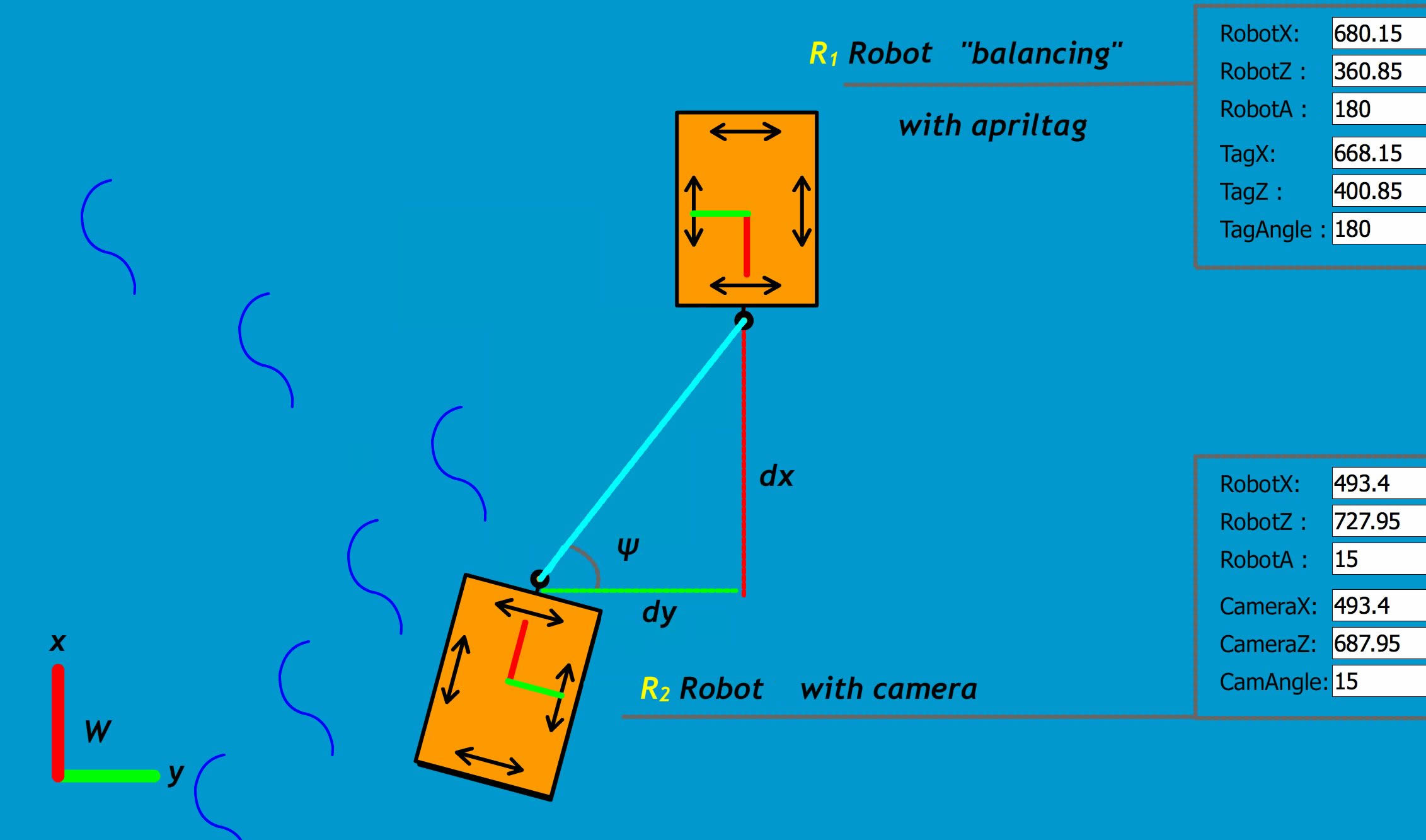}
	\end{center}
	\caption{2D working space for latching, $d_y$ lateral distance, $d_x$ longitudinal distance and $\psi$ angle between the entities.}
	\label{fig:simbase}
\end{figure}

The robot's localization is based on lidar with NDT matching \cite{7995900} and has an accuracy of $\pm 100mm$ in open spaces. However, this precision is not enough for putting together the boats on open water, since an accuracy of $\pm 40mm$ is required for performing the latching between the robots.
The robotic boats simplify the 3D positioning to a 2D challenge. Since, the entities are floating at similar levels above the water and the misalignment from the waves are compensated by the funnel, see Figure \ref{fig:simbase}.

Therefore, we integrated the AprilTags frameworks in the autonomous robotic boats for guiding one robot to a target, which can be a docking station or to another robot. 

The classic AprilTags framework is able to guide the robot in environments with low noise, while in noisy conditions, the $yaw$ angle varies several degrees and the robot is not able to reach the target, see Section \ref{classic}.
On the other hand, the AprilTags3D framework is able to overcome noises from the environment and precisely position the robot within the tolerances and perform the latching, see Section \ref{3dsec}.



\section{Experimental Results}

The challenge we want to solve with AprilTags3D is to have a reliable marker detection and high accuracy in pose estimation in highly reflective environments. In addition, we want to indirectly communicate the robot's state with only  the dynamic markers in a swarm robotics fashion. 

We performed three different experiments, in the first two tests we used classic printed AprilTag and our AprilTags3D to compare the results:
1) One robotic boat latch to a docking station in an indoors swimming pool with light reflections from the windows and head lamps. 
2) Instead of latching to a docking station, we set two robotic boats to latch outdoors on the open water. 
3) We test the indirect communication for swarm robotics with three autonomous robotic boats performing the $train$ $link$ formation with the dynamic AprilTags3D. Videos from the experiments available in \cite{luis}


\begin{figure}[t]
	\begin{center}
		\includegraphics[width=0.95\linewidth]{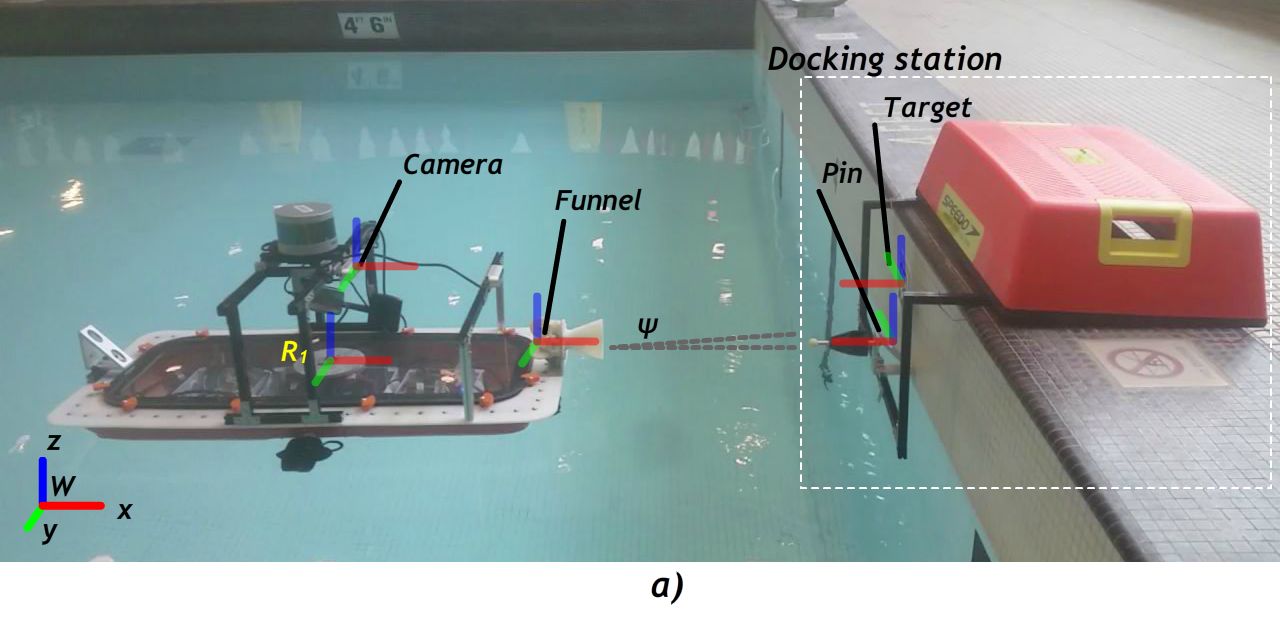}
		\includegraphics[width=0.95\linewidth]{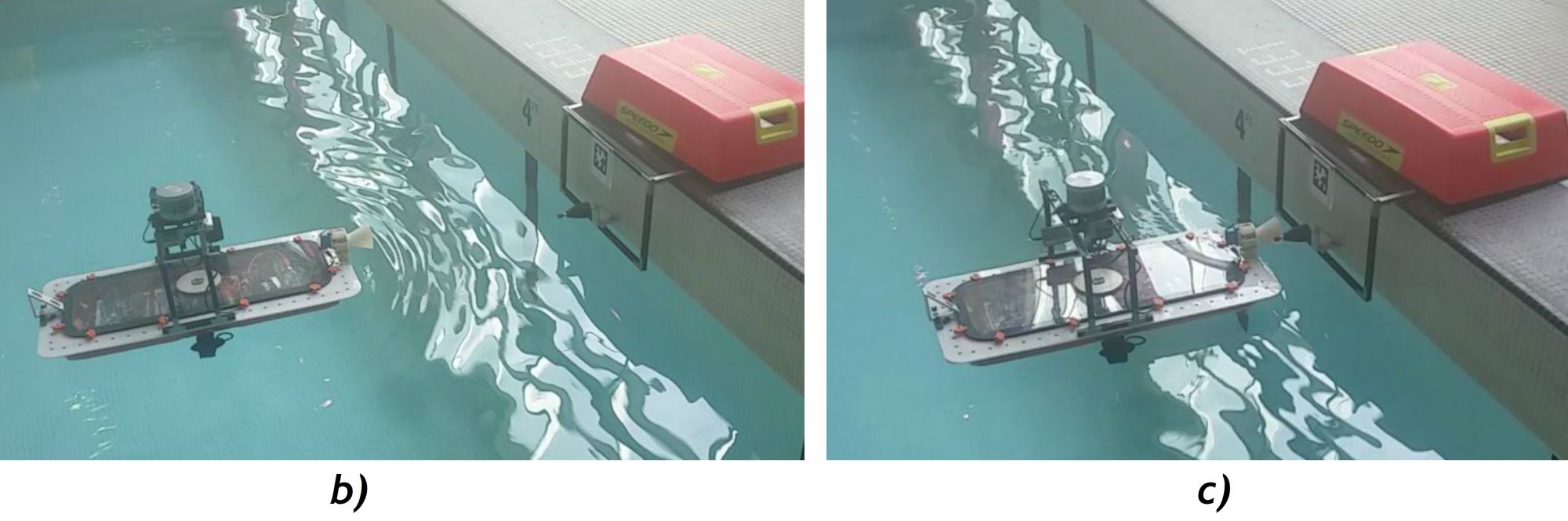}
		\includegraphics[width=0.99\linewidth]{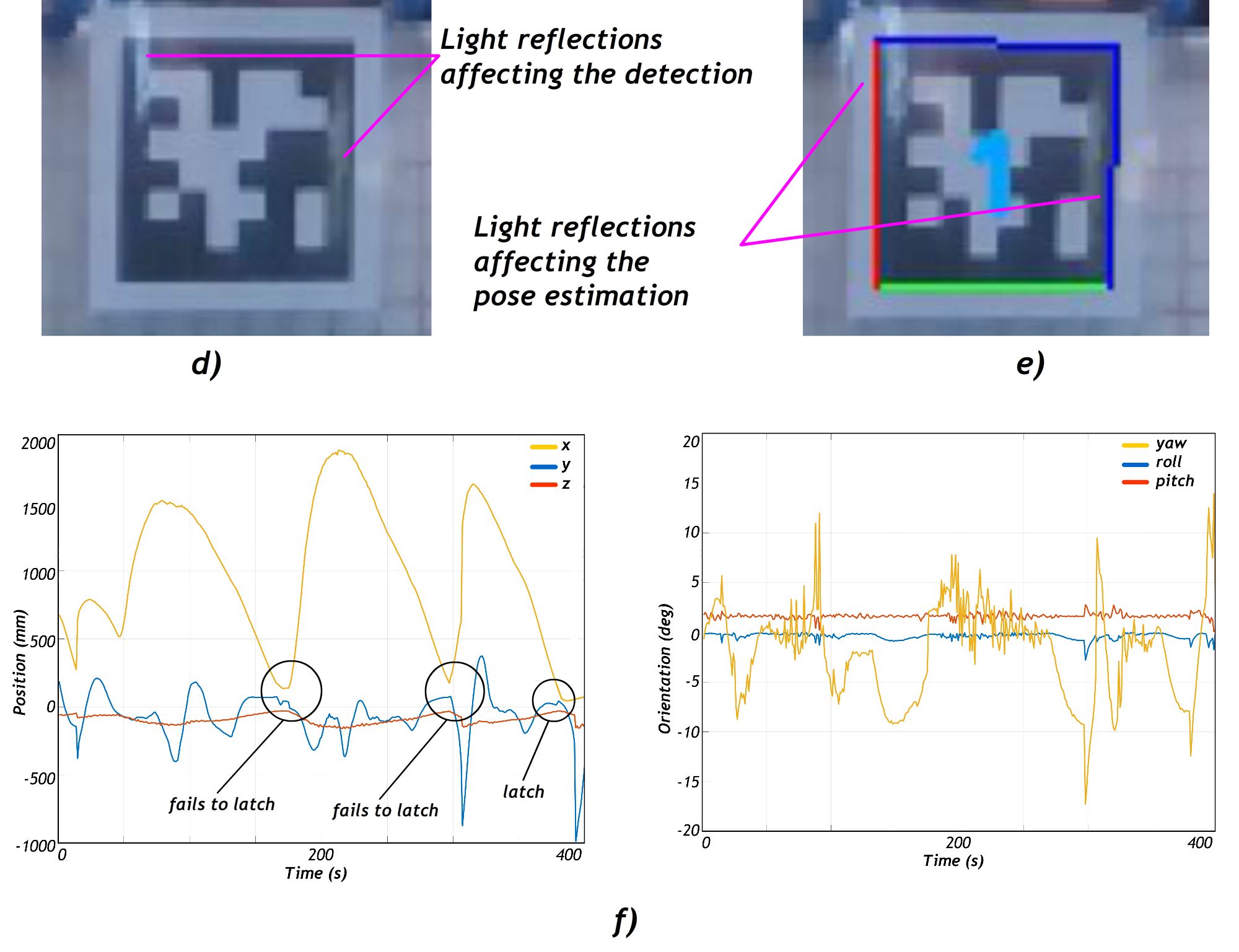}		
	\end{center}
	\caption{A) Autonomous robotic boat guided with classic AprilTags to latch to a docking station. The marker is used for detection and pose estimation of the docking station. B) The robot is initially set apart $1800mm$  from the tag in a direct view. C) The robot ultimately fails to dock due to the reflections on the target. D) The camera on the robot registered light reflections on the marker, which introduced extra white lines, affecting the detection. E) The camera on the robot registered light reflections resulting in a noisy pose estimation. F) Position and orientation of the robot when guided for latching.}
	\label{fig:tagswim}
\end{figure}

\subsection{Indoor test in swimming pool}
The setup consist of one autonomous robotic boat $R_1$ with a RGB camera and an actuated funnel, $R_1$ is located in a position $(R_{1_x},R_{1_y})$ in the swimming pool. $R_{1_z}$ is dismissed, since the controllable space is defined as a 2D plane on the water $(W_x,W_y)$.  
A docking station with a squared tag dimension $l=0.13m$ and a pin for latching is located at $(T_x,T_y,T_z)$.

The robot's camera can detect and calculate the pose of the marker, which is located on the docking station a couple of meters apart. In case, the robot fails to latch due to noise in the pose estimation readings, it  executes an auto-recovery algorithm to reposition itself to retry the docking action \cite{latching}. 

A successful latching is achieved when $d_x<10mm$, $d_y<\pm40mm$ and the $yaw$ angle is $<\pm27.5^\circ$.

\begin{figure}[t]
	\begin{center}
		\includegraphics[width=0.99\linewidth]{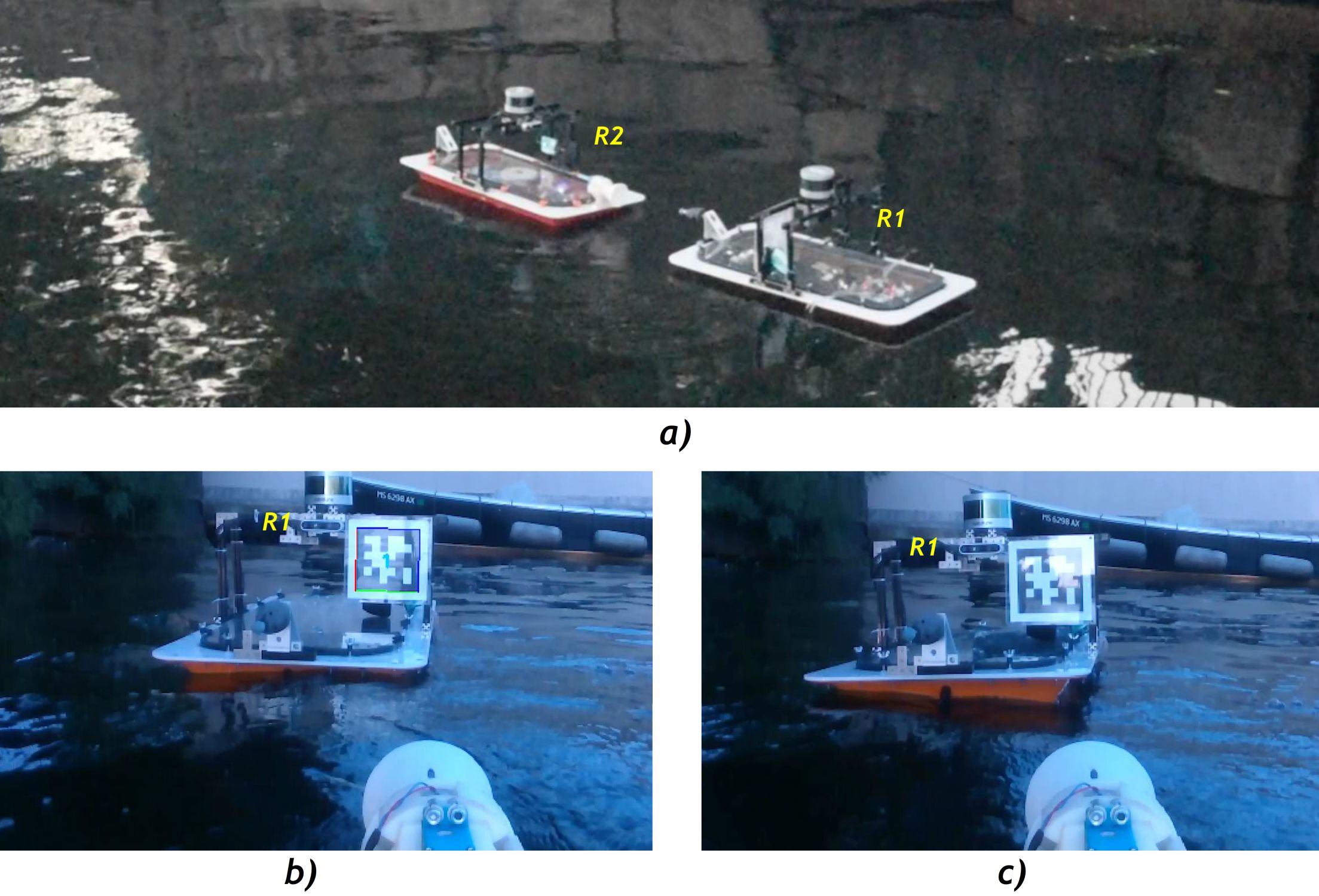}
		\includegraphics[width=0.99\linewidth]{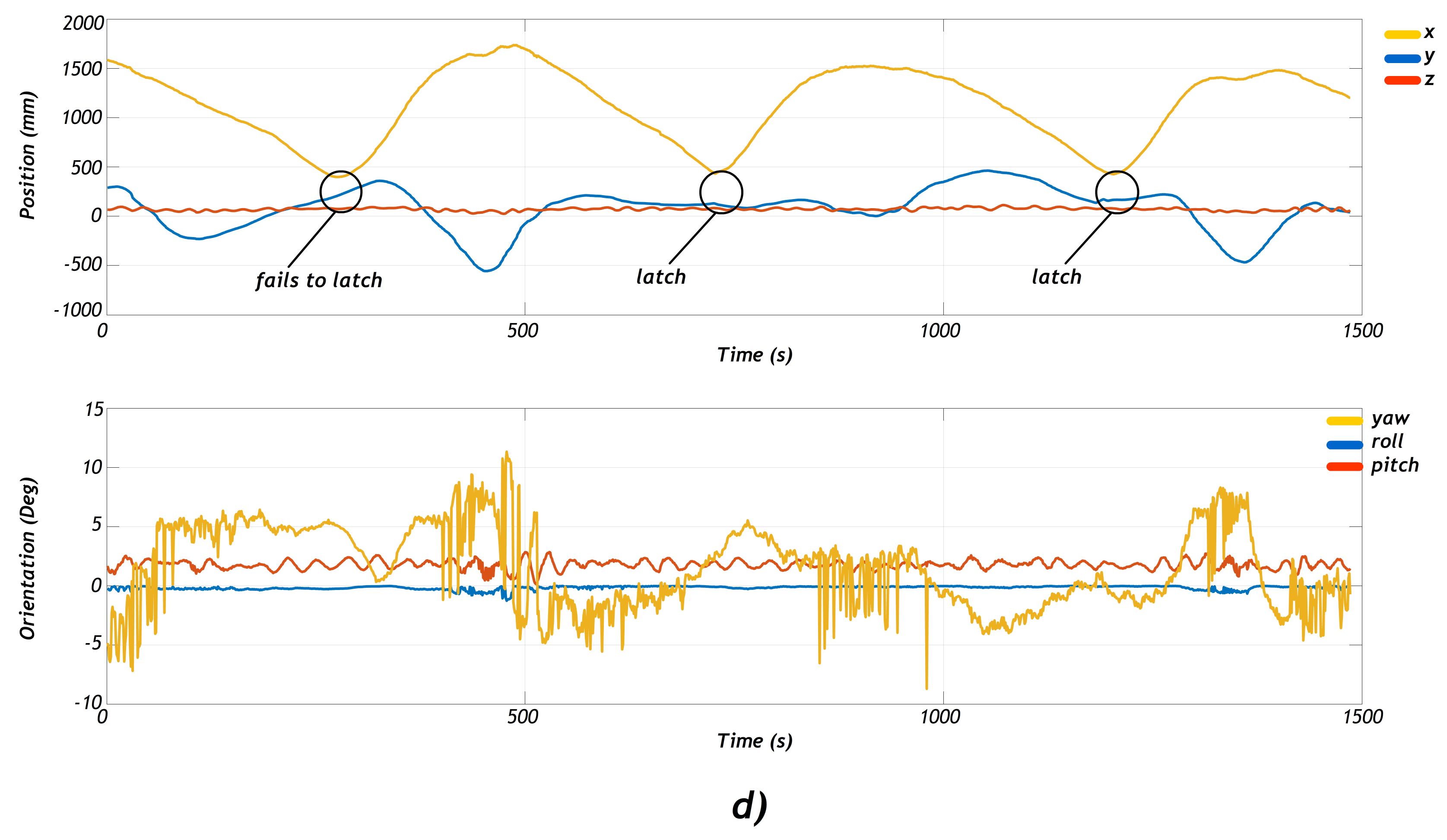}
	\end{center}
	\caption{A) Autonomous robotic boat guided with classic printed AprilTags to latch another floating robot in open water. B) The marker is detected with noisy pose estimation from the sky/sun light reflections. C) The marker is not detected from the light reflection that covers one quarter of the marker size. D) Position and orientation from the experiment. In this sequence, the robot initially fails to latch and then latches a couple of times.} 
	\label{fig:tagsun}
\end{figure}

\subsubsection{Indoor test with classic printed AprilTag}\label{classic}
The robot is initially set $1800mm$ away from the docking station in a direct line, meaning that the only variable to minimize for latching is the distance in the $X-axis$, $d_x=1800mm$, $d_y=0$ with $roll$, $pitch$ and $yaw$ $RPY=0$.

In the swimming pool we registered noisy pose estimation, the $yaw$ angle readings varied from $0.1^\circ$ to $10^\circ$. Making the robot to fail when latching, as the robot needed to try a couple of times before a successful latching. 
Figure \ref{fig:tagswim} shows the experimental setup and the sources of error from the light reflections on the marker. 



\subsubsection{Indoor test with dynamic AprilTag3D}\label{3dsec}
We integrated AprilTags3D with a couple of markers on the docking station. The $Tag_{leader}$ is set to replace the
printed version with position $(T_x,T_y,T_z)$, while the $Tag_{follower}$ is connected to the $Tag_{leader}$ on the $Z - axis$ and rotated $10^\circ$ on $yaw$, see Figure \ref{fig:roboat}$a$. 

The results show an improved detection and pose estimation, enabling the robotic boat to find and latch to the pin $99\%$ of the time.
Table \ref{tab:alles} compares the detection rate and pose estimation error with both, the classic AprilTags and  the AprilTags3D.

\begin{table}[]
	\begin{center}
		\begin{tabular}{|l|l|l|}
			\hline
			& \textit{Tag detection \% } & \textit{Yaw angle error (Deg)} \\ \hline
			\textbf{Indoors}  &                                   &                                                                \\ \hline
			AprilTags & 85\%                              & $\pm4^\circ$                                                             \\ \hline
			AprilTags3D       & 99\%                              & $< \pm1^\circ$                                                        \\ \hline
			\textbf{Outdoors} &                                   &                                                              \\ \hline
			AprilTags & 60\%                              & $\pm6^\circ$                                                       \\ \hline
			AprilTags3D       & 95\%                              & $< \pm1^\circ$                                                           \\ \hline
		\end{tabular}
	\end{center}
	\caption{Tag detection percentage and pose estimation error from testing indoors and outdoors with both, the classic AprilTags and AprilTags3D (a couple of dynamic markers).} 
	\label{tab:alles}
\end{table}

\subsection{Outdoor test in River}
The setup consist of two autonomous robotic boats performing a latching action on the river. The robot carrying the tag  $R_1$ tries to maintain its position on $(R_{1_x},R_{1_y})$ from its lidar sensor, with an error in position in the range of $\pm100mm$. While, robot $R_2$ tries to detect the tag, compute the pose estimation and latch to the "balancing $R_1$ robot".
The robot  $R_1$ integrates the tag and the pin, and the robot $R_2$ integrates the camera with the actuated funnel. 

Initially, $R_2$ is set $2000mm$ away from $R_1$ in a direct line, similar to the previous experiments in the swimming pool. 
In the same way, if the robot fails to latch at the first try, it will retry again autonomously until it latches. 

In this experiments, a successful latch is achieved when $d_y<\pm40mm$, $yaw$ angle is $<\pm27.5^\circ$ and $d_x<500mm$. Since, the tag is mounted inside the robot, see Figure \ref{fig:roboat}$b$.

\begin{figure}[t]
	\begin{center}
		\includegraphics[width=0.9\linewidth]{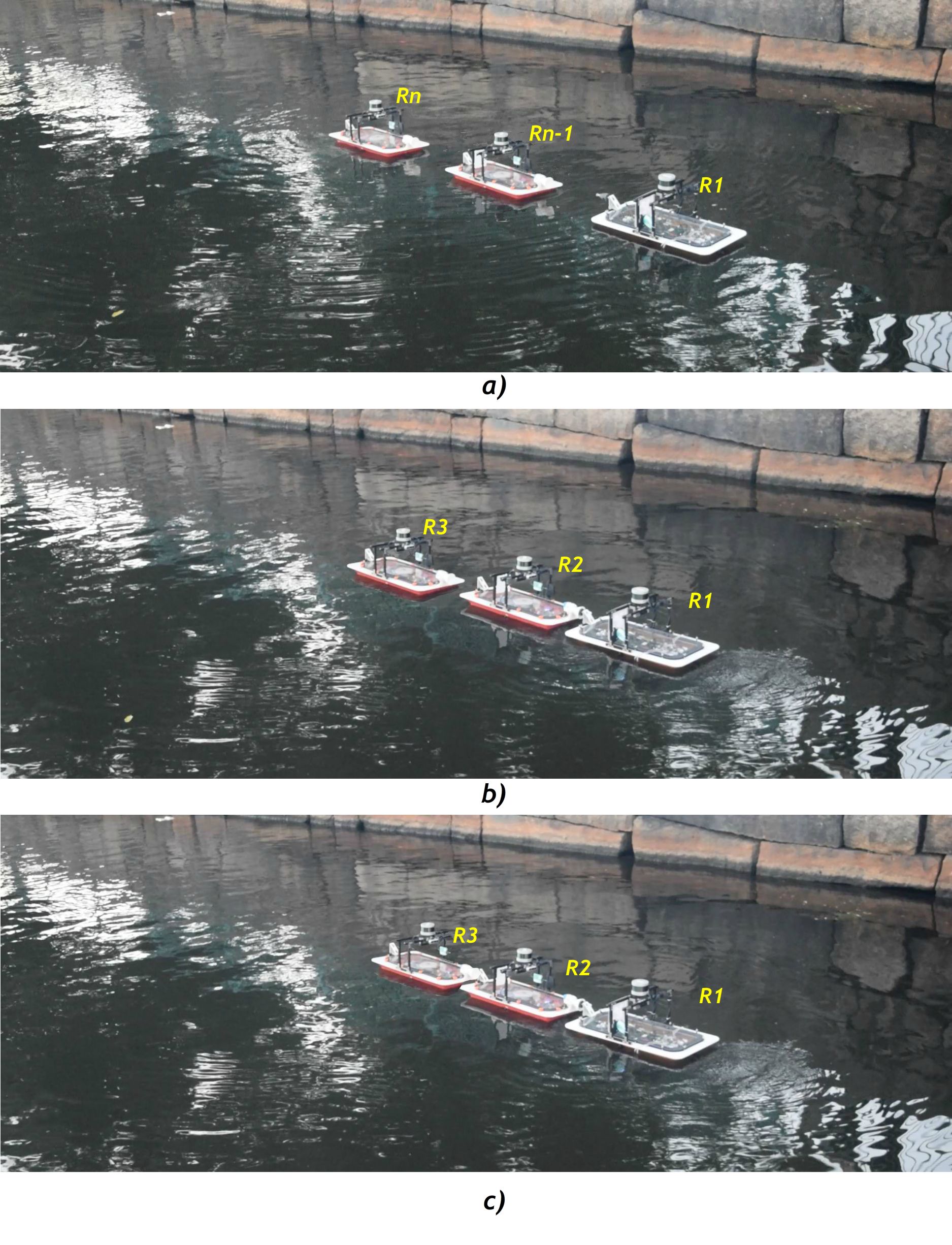}
	\end{center}
	\caption{A) The robotic boats are set in $stand$ $by$ configuration spacing each other $1m$, waiting for an order from  $R_1$. In a generic formulation for $n$ robots: $R_n$ can see $R_{n-1}$, $R_{n-1}$ can see $R_{n-2}$ and so on, until $R_1$. B) The leader robot $R_1$ changes its tag to let know the next robot $R_2$ to configure in $train$ $link$ and to pass the message to the next robot in line. $R_2$ latches $R_1$ and changes its tag to let know $R_3$ that it is latched and implicitly communicates the type of configuration. C) $R_3$ latches accordingly.}
	\label{fig:swarmreal}
\end{figure}

\subsubsection{Outdoor test with classic printed AprilTag}

In order to latch the robots, robot $R_2$ is guided by visual cues to a position $(x,y)$. This coordinate must be reach with an $error < \pm 40mm$ for a successful latching of robot $R_1$. In this challenging task, the robots are not stable on the ground, they are floating on the river, trying to maintain their positions, overcoming waves, wind and water currents.  

From the camera point of view it is a big challenge to process the 6DOF from the tags located in another boat when it moves with the waves at a degree that the $pitch$ and $roll$ angles vary a few degrees. Further, we are only relying on RGB sensors to perform the latching on the boats.

On the river, the main problem with the classic markers was the detection. The marker was  detected only $60\%$ of the time, due to the sunlight reflections, see Figure \ref{fig:tagsun}.  
The reason for this low detection rate were the waves, since these created oscillations on both robots, changing their position and inclination, creating multiple random reflections on the tag. In addition, the light reflections from cars on the avenue and boats nearby created an extremely challenging condition for latching the robots. 
In this real environment $R_2$ required in average five attempts before a successful latching.

\subsubsection{Outdoor test with dynamic AprilTag3D}
We integrated a couple of tags on the robot $R_1$. The $Tag_{leader}$ replaced the printed tag, with position $(R_{1T_x},R_{1T_y}, R_{1T_z})$, and the $Tag_{follower}$ was set similar to the indoor experiments, see Figure \ref{fig:roboat}$b$. 

The results revealed a consistent tag detection, detecting at least one tag $95\%$ of time. The framework shows a minimization in the environmental noises and an improved detection rate, see Table \ref{tab:alles}. 

\subsection{Dynamic AprilTag3D for swarm robotics}\label{swarms}
In our framework for joining the robots, the latching state of the robot can be registered. When a robot latches, the actuated funnel closes a socket, trapping the bearing stud from another robot or from the docking station. In this sense, the robot performing the latching action can communicate the state of its actuated funnel by changing its tag dynamically, on its LCD screen. 

Since each robot is an independent entity and each robot speaks the same language, meaning they understand a set of common tags. Then, they can communicate in a swarm robotic fashion with the constraint  that the robots must be able to see and detect the tag on another robot.

\begin{figure}[b]
	\begin{center}
		\includegraphics[width=0.9\linewidth]{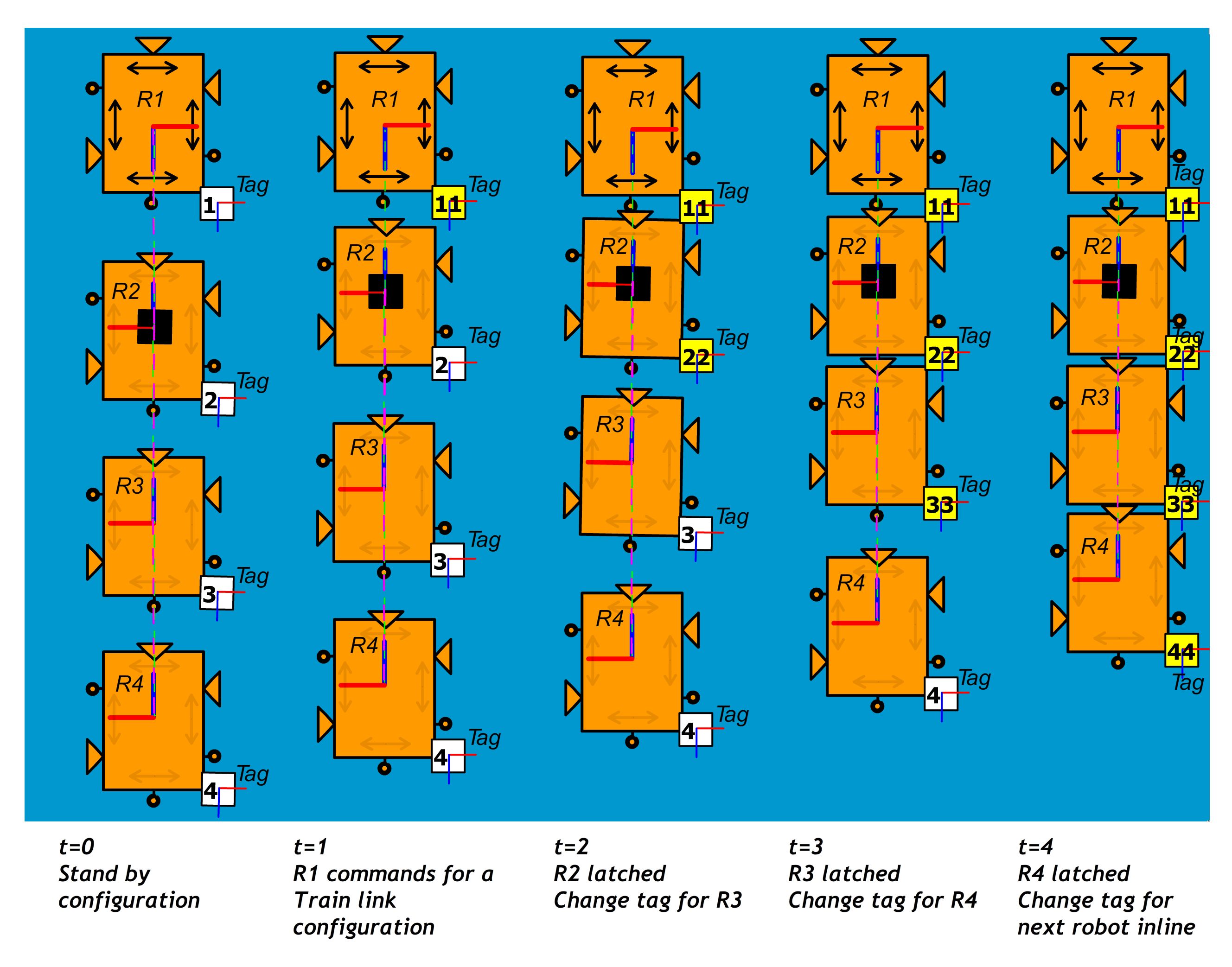}
	\end{center}
	\caption{Camera - tag communication for robot formation. At $t=0$, the robots are spaced $1m$ each other waiting for a change in the tag on the robot in front on them. At $t=1$, the robot leader $R_1$  commands the swarm robots in a $train$ $link$ configuration. At $t=2$, $R_2$ computes the pose estimation of $R_1$, latches and changes its tag status to inform the next robot $R_3$ to perform a latching. At $t=3$, $R_3$ calculates the pose of $R_2$, latches and changes its tag status to inform $R_4$ to perform a latching. 
	} 
	\label{fig:swarmsim}
\end{figure}

This framework works for multiple case scenarios, Figure \ref{fig:swarmreal} shows the robotic boats performing a $train$ $link$ formation. 
Moreover, this case scenario can be scale to $N$ robots with minimal complexity. The framework can be seen as the robots playing the telephone game. One robot passes a message to another and this one passes to the next ad so on. The passing of message can encode a robotic formation, a position and/or a change of state, see Figure \ref{fig:swarmsim}.

\section{Conclusion}
We presented a novel methodology of squared fiducial markers called AprilTags3D which improves the detection and pose estimation of the classic AprilTags in field robotics in real environments.
 
We performed multiple tests indoors and outdoors with robotic boats that autonomously latch to a docking station and to another robot with the camera - target principle. 
The experimental results using both methodologies, revealed that our method improves the detection and pose estimation in real case scenarios. Specially, in highly reflective environment, such as aquatic environments inside the city (rivers and lakes) with sunlight reflections from cars, building and other boats. 

In the experiments we simplified the 3D space to a 2D plane on the water surface for guiding the robotic boats. Nevertheless, the proposed AprilTags3D framework can be integrated in drones, submarines or space robotics required to navigate in 3D space. 


Also, we presented a novel concept to indirectly communicate robots in a swarm by only changing their dynamic tag depending on its state. This simple rather powerful framework can be scaled to $N$ robots with minimal complexity for robot formations. We showed how three robotic boats "talk" to each other to perform robotic formations on the river.


{\small
\bibliographystyle{ieee}
\bibliography{april}
}

\end{document}